\documentclass[letterpaper]{article} 
\usepackage{aaai25}  
\usepackage{times}  
\usepackage{helvet}  
\usepackage{courier}  
\usepackage[hyphens]{url}  
\usepackage{graphicx} 
\usepackage{natbib}  
\usepackage{caption} 
\usepackage{algorithm}
\usepackage{algorithmic}
\usepackage[table]{xcolor}  
\usepackage{multirow}
\usepackage{newfloat}
\usepackage{listings}
\DeclareCaptionStyle{ruled}{labelfont=normalfont,labelsep=colon,strut=off} 
\floatstyle{ruled}
\newfloat{listing}{tb}{lst}{}
\floatname{listing}{Listing}
\title{RCTrans: Radar-Camera Transformer \\ via Radar Densifier and Sequential Decoder for 3D Object Detection}
\author{
    Yiheng Li\textsuperscript{\rm 1,}\textsuperscript{\rm 2,}\equalcontrib ,
    Yang Yang\textsuperscript{\rm 1,}\textsuperscript{\rm 2,}\equalcontrib\textsuperscript{\rm },
    Zhen Lei \textsuperscript{\rm 1,}\textsuperscript{\rm 2,}\textsuperscript{\rm 3,}\footnote{Corresponding author.}
}
\affiliations{

    \textsuperscript{\rm 1} State Key Laboratory of Multimodal Artificial Intelligence Systems (MAIS),\\ Institute of Automation, Chinese Academy of Sciences\\
    \textsuperscript{\rm 2} School of Artificial Intelligence, University of Chinese Academy of Sciences\\
    \textsuperscript{\rm 3}Centre for Artificial Intelligence and Robotics, Hong Kong Institute of Science \& Innovation,\\
    Chinese Academy of Sciences

    liyiheng2024@ia.ac.cn, yang.yang@nlpr.ia.ac.cn, zhen.lei@ia.ac.cn
}

\usepackage{bibentry}

\begin{document}

\maketitle

\begin{abstract}
In radar-camera 3D object detection, the radar point clouds are sparse and noisy, which causes difficulties in fusing camera and radar modalities. To solve this, we introduce a novel query-based detection method named Radar-Camera Transformer (RCTrans). Specifically, we first design a Radar Dense Encoder to enrich the sparse valid radar tokens, and then concatenate them with the image tokens. By doing this, we can fully explore the 3D information of each interest region and reduce the interference of empty tokens during the fusing stage. We then design a Pruning Sequential Decoder to predict 3D boxes based on the obtained tokens and random initialized queries. To alleviate the effect of elevation ambiguity in radar point clouds, we gradually locate the position of the object via a sequential fusion structure. It helps to get more precise and flexible correspondences between tokens and queries. A pruning training strategy is adopted in the decoder, which can save much time during inference and inhibit queries from losing their distinctiveness. Extensive experiments on the large-scale nuScenes dataset prove the superiority of our method, and we also achieve new state-of-the-art radar-camera 3D detection results. Our implementation is available at https://github.com/liyih/RCTrans.

\end{abstract}

%

\section{Introduction}
3D object detection is an important perceptual task that can be widely applied in many fields. Existing high-precision detection algorithms often rely on the input of LiDAR\cite{yan2018second,yin2021center}. However, LiDAR is expensive and easily damaged \cite{lin2024rcbevdet}, which is not conducive to the commercialization of the algorithm. To reduce the practical cost, recent works begin to focus on using low-cost radars and multi-view cameras to improve detection performance. Camera sensors can provide rich texture information, while radar sensors can provide three-dimensional information that is not affected by different weather conditions \cite{kim2023craft}. 

However, the radar sensor has two inherent drawbacks, i.e., sparsity and noise. 
Firstly, a radar typically releases mmWave into the physical space, which hits the object and then is reflected to the receiver. Due to the lack of diffused reflection, only a small part of objects can be reflected to the radar receiver \cite{singh2023depth}. The number of non-empty radar pillars is only approximately 10\% of average numbers in LiDAR pillars \cite{bang2024radardistill}. Secondly, the radar used in autonomous driving (such as in nuScenes \cite{caesar2020nuscenes}) cannot obtain an accurate height of objects, which can bring the noise in azimuth (x) and depth (z) components \cite{singh2023depth}. In addition, some noise may also be led to the axes when the beam-width of the mmWave radar is too large \cite{singh2023depth}. The sparsity and noise of radar data will bring challenges to the cross-modality fusion. Specifically, sparsity causes too much invalid radar representation, while noise makes wrong or inaccurate alignments between camera and radar representation.

\begin{figure}
    \centering
    \includegraphics[scale=0.5]{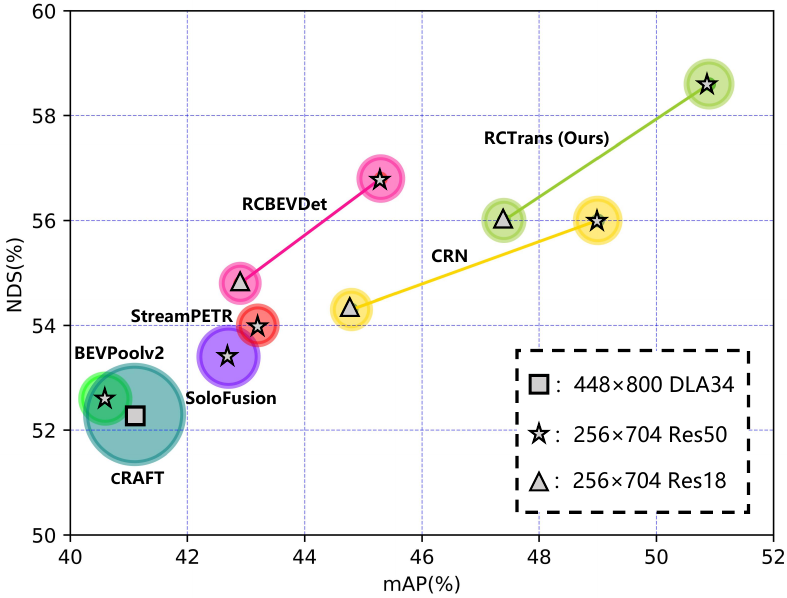}

    \caption{Comparison between RCTrans and other camera or radar-camera 3D detection methods. The size of the circle represents the time consumption of inference. All experiments are conducted on nuScenes \texttt{val} set and the speed is measured on a single NVIDIA RTX3090 GPU.}
    \label{pic1}
\vspace{-0.5cm}
\end{figure} 

As shown in Fig. \ref{pic1}, we compare RCTrans with existing models on nuScenes \texttt{val} set.
Previous methods may overlook the sparsity of the radar \cite{kim2023crn, kim2023craft}, or need the radar cross-section (RCS) as the prior \cite{lin2024rcbevdet}. They cannot adaptively fill every empty BEV grid. Moreover, the SOTA radar-camera solutions \cite{kim2023crn,lin2024rcbevdet} adopt bird’s eye view (BEV) \cite{huang2021bevdet} as the unified space to fuse features. Such kind of methods require strict correspondences between multi-modality features in BEV space, but it is difficult to establish precise correspondences for radar with positional offset. Inspired by the success of query-based 3D detection \cite{liu2022petr,yan2023cross}, we use the query-based paradigm to establish interaction between object queries and multi-modality tokens. A positional embedding is used to construct implicit associations, allowing the model to adaptively learn more flexible correspondences.


In this paper, we propose a novel Radar-Camera Transformer (RCTrans) framework via radar densifier and sequential
decoder to overcome the sparsity and noise of radar inputs. 
To be specific, we design a Radar Dense Encoder (RDE) to avoid blurring some detailed information while densifying every empty BEV grid.
It uses a downsample-then-upsample structure and obtains multi-scale information through a skip connection. Self-attention is adopted at the smallest BEV scale to adaptively fill each grid. 
Considering that information from different modalities can establish more flexible interaction with object queries.
we design the Pruning Sequential Decoder, which gradually fuses multi-modality features and locates the position of each object in a step-by-step way. 
In each layer, two separate transformers will be used to fuse the radar and image information, respectively. This approach allows the model to establish the connections between different modalities and object queries based on their distinct traits.
Thereafter, we predict the new position of queries and recompute position embedding after each decoder layer. In the next layer, we can use the updated position embedding to calculate more precise correspondences and obtain more accurate fusion results. A pruning training strategy is used to speed up the inference and prevent object queries from losing their distinctiveness.

We conduct extensive experiments on the large-scale autonomous driving dataset nuScenes \cite{caesar2020nuscenes}, which demonstrates the effectiveness of our proposed RCTrans. 
The ablation study shows that our designed  Radar Dense Encoder and Pruning Sequential Decoder both contribute to improving the performance. 
Our contributions are summarized as:
\begin{itemize}
    \item We introduce a new query-based 3D detection framework named Radar-Camera Transformer (RCTrans), which can solve the problems caused by the sparsity and noise of radar point clouds. 
    \item We first design the Radar Dense Encoder to densify the sparse radar features, and then propose the Pruning Sequential Decoder to make flexible alignment and effective fusion in the prediction process.
    \item We confirm the efficacy of RCTrans by achieving superior 3D detection performance in the nuScenes dataset with an acceptable speed. What's more, the ablation study proves the role of each proposed module.
\end{itemize}

\section{Related Work}

\subsection{BEV-Based 3D Object Detection}
Bird's-eye-view (BEV) is a dense 2D representation, where each grid corresponds to a certain region in the scene \cite{mao20233d}. LSS \cite{philion2020lift} innovatively solves the transformation problem from multi-view cameras to BEV features. BEVDet \cite{huang2021bevdet} is a pioneer in applying BEV to 3D object detection. The major difficulty of building BEV lies in inaccurate depth estimation, and BEVDepth \cite{li2023bevdepth} solves this bottleneck by using additional depth supervision. In order to predict more accurate speed and reduce the missed detection of objects, more and more works try to incorporate temporal information.
BEVFormer \cite{li2022bevformer} is the first to introduce sequential temporal modeling in multi-view 3D object detection. SOLOFusion \cite{park2022time} proposes a stream video paradigm to make long-term temporal modeling. BEV can also serve as a unified representation space for the fusion of multi-modality features. BEVFusion \cite{liu2023bevfusion,liang2022bevfusion} is a pioneer in choosing BEV as a unified representation space for fusion. BEV can provide explicit correspondence for different modalities, which is extremely convenient for fusion. However, alignment errors between different modal BEVs can affect the fusion effect. Some approaches \cite{song2024graphbev, lin2024rcbevdet} try to solve this issue via a deformable module.

\subsection{Query-Based 3D Object Detection}
The query-based method is another important paradigm in 3D object detection. In these methods, object queries typically interact with the tokens of input data. Inspired by the successful query-based methods in 2D detection \cite{carion2020end}, DETR3D \cite{wang2022detr3d} introduces a set of points to be the references, each of them represents an object query. Later, researchers begin to focus on how to effectively extract features from multi-view images to the object queries.
PETR \cite{liu2022petr} proposes 3D position-aware position embedding for image features in the LiDAR coordinate. CAPE \cite{xiong2023cape} constructs the position embedding in camera coordinates to reduce the influence of variant camera extrinsic. StreamPETR \cite{wang2023exploring} proposes a query-based long-temporal fusion algorithm that can transfer historical information from objects to the current frame through object queries. PETRv2 \cite{liu2023petrv2} aggregates short-temporal information by transforming the previous 3D coordinates to the current coordinate system. Some methods also explore the query-based multi-modality fusion paradigm. Futr3d \cite{chen2023futr3d} introduces the first unified end-to-end sensor fusion framework in 3D detection. CMT \cite{yan2023cross} proposes a 3D position embedding to fuse multi-modality tokens.

\begin{figure*} [t]
    \centering
    \includegraphics[scale=0.53]{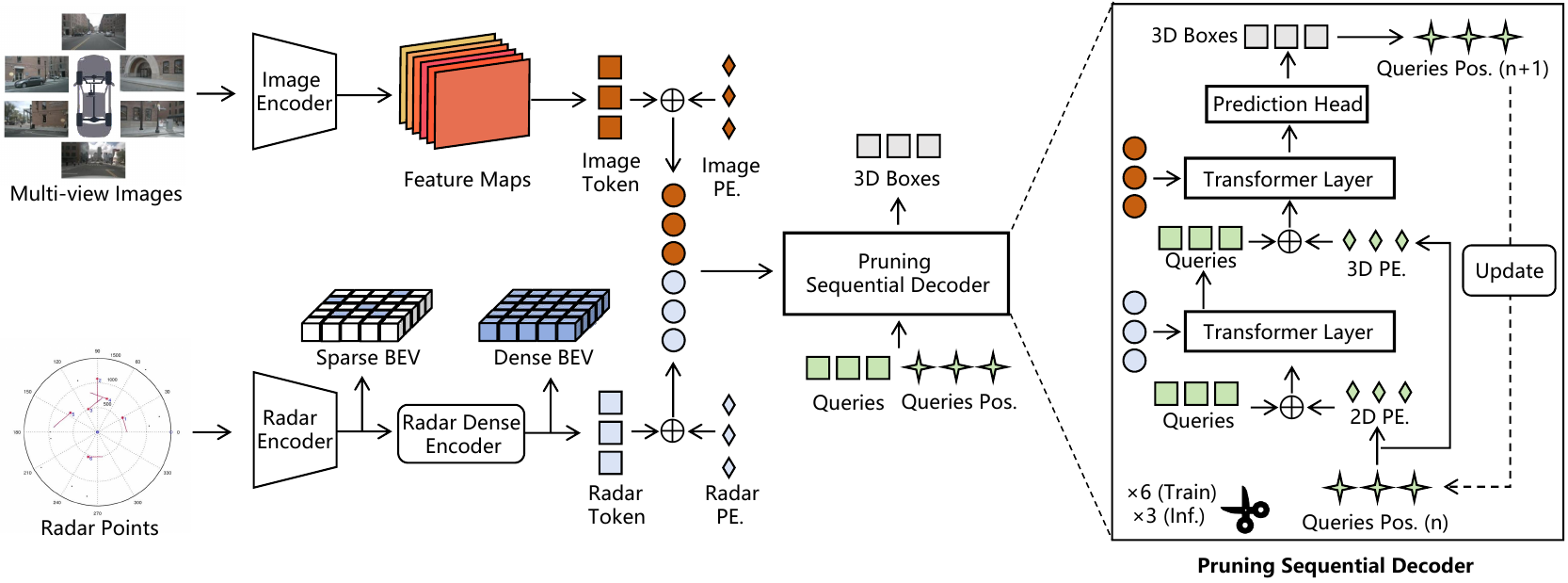}
    
    \caption{The overall architecture of Radar-Camera Transformer. RCTrans can be divided into two key components: (1) Token generator which extracts multi-modality tokens and sums them with corresponding position embedding and (2) Pruning Sequential Decoder which updates the randomly initialized queries and queries position in each layer and makes more precise fusion in a step-by-step way. The final predicted results are the outputs of the last layer after pruning.} 
    \label{fig2}
\end{figure*}

\subsection{Radar-Camera 3D Object Detection}

To explore low-cost 3D perception solutions, more and more methods have begun to focus on using radar sensors. RadarNet \cite{yang2020radarnet} utilizes radar information to complement other sensors in the form of Doppler velocity.  
Based on the frustum-based association method, CenterFusion \cite{nabati2021centerfusion} accurately associates radar detections to objects in the image. What's more, it adopts a middle-fusion approach to complement the image features via radar feature maps. CRAFT \cite{kim2023craft} introduces the Spatio-Contextual Fusion Transformer to conduct radar-camera detection. CRN \cite{kim2023crn} transforms the image feature into BEV space with the help of radar points. RADIANT \cite{long2023radiant} predicts the object center offsets and associates these offsets with the image detector. It then obtains accurate depth estimates for object detection and enhances camera features with estimated depth. RCFusion \cite{zheng2023rcfusion} proposes a weighted fusion module to fuse radar and camera BEV features.
RCBEVDet \cite{lin2024rcbevdet} fuses and predicts in the BEV space and proposes the RadarBEVNet to better extract radar features.

In contrast, we introduce a query-based fusion method named RCTrans, which focuses on how to solve the inherent drawbacks of the radar sensor and achieves high results.

\section{Method}

The pipeline of our proposed Radar-Camera Transformer (RCTrans) is shown in Fig. \ref{fig2}. Firstly, two parallel branches are used to extract multi-modality tokens. In the radar branch, we will use the Radar Dense Encoder to densify the valid radar features. Then, position embeddings of multi-modality tokens are generated and added to the tokens. Finally, the randomly initialized queries will be sent to the Pruning Sequential Decoder along with the tokens to predict 3D boxes. 
The entire training is an end-to-end process and does not require freezing any parameters.

\subsection{Token Generator}

For the camera branch, given the images $I \in R^{N_i \times H\times W\times 3}$ from $N_i$ views where $H$ and $W$ are the size of the image, the image encoder is used to encode the images into feature maps $F \in R^{N_i \times \frac{H}{16}\times \frac{W}{16} \times C}$. Then, the $F$ is flattened into the image tokens. For the radar branch, we follow Futr3d \cite{chen2023futr3d} to use the 3D coordinate, compensated velocities, and the time offset as the input channel of the radar points $P \in R^{N_r \times 5}$, where $N_r$ is the number of the points. After that, we use MLP to obtain the features of each pillar and scatter them to the corresponding BEV grids. Due to the sparsity of radar points, more than 90\% BEV grids are empty, which may hurt the following detection results. To solve it, we introduce a Radar Dense Encoder to adaptively fill every grid while also keeping multi-scale information.

\begin{figure*}[t]
    \centering
    \includegraphics[height=3cm, width=17.5cm]{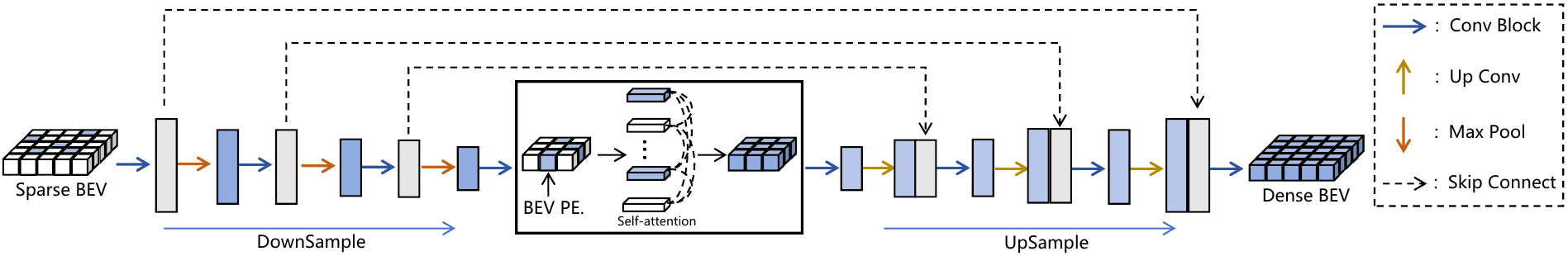}
    \caption{The overall architecture of Radar Dense Encoder. RDE uses a downsample-then-upsample architecture. Self-attention is used to fill the empty grids in the minimal resolution. Skip connection is used in the upsampling process.} 
    \label{fig3}
\end{figure*}

\textbf{Radar Dense Encoder.} One naive solution is to directly increase the number of the BEV encoder layers. However, this may smooth the small objects in the background. Moreover, commonly used BEV encoders, such as SECOND \cite{yan2018second}, are mainly designed for LiDAR and not entirely suitable for radar inputs. RCBEVDet \cite{lin2024rcbevdet} proposes to use the radar cross-section (RCS) to determine the scattering radius of radar points. However, the scattered results rely on the quality of RCS and the number of scattered features in each BEV grid is different. We aim to find a network with a simple structure, that can adaptively fill each BEV gird and aggregate multi-scale information to facilitate detection of objects with different sizes.

The simplest method to adaptively fill each BEV grid is to use the global self-attention mechanism. However, due to the large number of BEV grids, this method consumes a long time. To solve it, we first downsample the original BEV features and interact with them in a minimal resolution. At the same time, the downsampling process can greatly reduce the number of invalid grids, thereby improving the quality of features obtained after interaction. Inspired by U-net \cite{ronneberger2015u}, we connect features of different receptive field sizes at the same BEV resolution, which effectively preserves the features of objects of different sizes. The architecture of the Radar Dense Encoder (RDE) is shown in Fig. \ref{fig3}. Specifically, given the radar BEV feature $B \in R^{H_r\times W_r\times C_r}$, where $H_r$ and $W_r$ are the size of the BEV map, we first extract multi-level downsampled features $B_d = \{ B_i \in R^{\frac{H_r}{2^i}\times \frac{W_r}{2^i}\times C_i}, i=1,2,3\}$. Then, we conduct self-attention in $B_3$ and obtain adaptively filled BEV map $B_f$. During this process, each grid is added by the 2D position embedding. After that, we upsample the $B_f$ to the same size as the original $B$. Skip connection is used to fuse the upsampled BEV map and the same size features of $B_d$ in the upsampling process.

\textbf{Position Embedding.} For image position embedding, we use the 3D position embedding proposed in PETR \cite{liu2022petr}. Given a image token $T_i$, a series points $p(u,v)=\{ p_i(u,v) = (u\times d_i, v\times d_i, d_i, 1), i = 1, 2, ..., d \}$ is pre-defined in the camera frustum space. Here, $u$ and $v$ are the indices of the token in image space, and $d$ is the number of points along the depth axis. After that, the image position embedding is calculated via Eq. \ref{eq1}. 
\begin{equation}\label{eq1}
PE_{im} = \Phi_{im}(Kp(u,v)),
\end{equation}
where $K$ is the transformation matrix that transforms camera frustum to 3D world space. $\Phi_{im}(.)$ is the MLP function. As the radar can not obtain precise height information of objects, for radar position embedding, we use the 2D BEV embedding, ignoring the height information of the BEV grid. The radar position embedding is calculated via Eq. \ref{eq2}. 
\begin{equation}\label{eq2}
PE_{ra} =\Phi_{ra}(\Psi(h,w)),
\end{equation}
where $(h, w)$ is the 2D coordinate of BEV grid, $\Phi_{ra}(.)$ is the MLP function, and $\Psi(.)$ is the sine-cosine function. Through positional embedding, information from different modalities can be implicitly aligned with object queries in 3D space.

\subsection{Pruning Sequential Decoder}
The Pruning Sequential Decoder consists of multiple layers, 
in each layer, we fuse the image tokens with object queries through $PE_{im}$ and $PE_{3d}$, and fuse the radar tokens with object queries through $PE_{ra}$ and $PE_{2d}$ due to the inaccurate height information of radars.
Following previous query-based methods \cite{liu2022petr, wang2022anchor}, we first initialize $n$ learnable references $R = \{ r_i = (r_{xi}, r_{yi}, r_{zi}), i=1,...,n \}$ in 3D scene to be the position of the object queries, which uniformly distribute between $[0, 1]$. The query feature $F_q$ corresponding to each reference is initialized as a vector of all zeros. 

\textbf{Query Feature Update.} 
The query positions $R$ are first projected
into radar and image space. The projected query positions $R_{ra}$ in radar space can be obtained by Eq. \ref{eq3}.
\begin{equation}\label{eq3}
\left\{  
    \begin{array}{lr} 
        r_{xi}^{'} = r_{xi}\times(x_{max} - x_{min}) + x_{min}\\ 
        r_{yi}^{'} = r_{yi}\times(y_{max} - y_{min}) + y_{min}\\ 
        r_{zi}^{'} = r_{zi}\times(z_{max} - z_{min}) + z_{min},\\
    \end{array}
\right.
\end{equation}
where $\{[x_{max}, x_{min}], [y_{max}, y_{min}], [z_{max}, z_{min}]\}$ is the range of valid 3D world space. The projected query positions $R_{im}$ in image space can be obtained by Eq. \ref{eq4}.
\begin{equation}\label{eq4}
R_{im} = K^{-1}R_{ra},
\end{equation}
where $K$ establishes the transformation from camera frustum to 3D world space. Following CMT \cite{yan2023cross}, we adopt the shared encoder used in image/radar position embedding generation to encode 3D/2D position embedding of queries. The $PE_{3d}$ and $PE_{2d}$ are obtained by Eq. \ref{eq5}.
\begin{equation}\label{eq5}
PE_{3d} = \Phi_{im}(R_{im}), PE_{2d} = \Phi_{ra}(\Psi(R_{ra}))
\end{equation}
Different from the common decoder used for multi-modality tokens which concatenates them and uses a large layer to fuse them together, we propose to split this large layer into two small ones and fuse the multi-modality tokens separately. We name it sequential structure. The updated queries are calculated by Eq. \ref{eq6}.
\begin{equation}\label{eq6}
F_{q}^{n+1} = f_2(f_1(F_{q}^{n}+PE_{2d}, T_{r})+PE_{3d}, T_{i}),
\end{equation}
where $T_{r}$ and $T_{i}$ represent radar and image tokens. $f_{1}(.,.)$ and $f_{2}(.,.)$ represent the transformer layer.

\textbf{Query Position Update.} At the end of each decoder layer, we will predict the positions of the queries. In the next layer, new position embedding will be generated based on the updated positions. Given the updated query $F_{q}^{n+1}$, we predict the offset $\Delta R$ of the query positions, and the updated position $R_{n+1}$ can be calculated by $R_{n}+\Delta R$.

\textbf{Pruning Training Strategy.} In each layer, we use a sequential structure to fuse multi-modality information, which results in using twice as many transformer layers as traditional decoders. This can cause additional inference time. What's more, as we update the positions of queries after each decoder layer, some object queries may gradually be located in the same region (please refer to visualization in appendix) and lose the feature distinctiveness. This causes information from some regions to be overlooked, and the attention mechanism may fail to learn effective representation learning concepts which prevents the model from achieving the expected performance improvement \cite{zhou2021deepvit}.
To this end, we propose a pruning training strategy, which uses 6 layers of decoders during training and only 3 layers of decoders during inference. 
\begin{table*}[!t]
\begin{center}
\begin{tabular}{l|c|c|c|cc|c}
    \hline
    Method & Input & Image Backbone & Image Size & NDS (\%) $\uparrow$ & mAP (\%) $\uparrow$ & Latency (ms) $\downarrow$ \\
    \hline
    CRN (\citeyear{kim2023crn}) & RC& ResNet18 & $256\times704$ & 54.3& 44.8& 35.8 \\
    RCBEVDet (\citeyear{lin2024rcbevdet}) & RC & ResNet18 &  $256\times704$ & 54.8 & 42.9 &\textbf{35.3}\\
    \rowcolor{gray!20} RCTrans(ours) & RC & ResNet18 &  $256\times704$ &56.0&47.4&39.8\\
    \hline
    
    BEVDepth (\citeyear{li2023bevdepth})
    & C & ResNet50 & $256\times704$ & 47.5 & 35.1 &86.2\\
    BEVPoolv2 (\citeyear{huang2022bevpoolv2})
    & C & ResNet50 & $256\times704$& 52.6 & 40.6 & 60.2\\
    SOLOFusion (\citeyear{park2022time})
    & C & ResNet50 & $256\times704$ & 53.4 & 42.7 &87.7\\
    StreamPETR (\citeyear{wang2023exploring})& C & ResNet50 & $256\times704$ & 54.0 & 43.2 &36.9\\
    CRN (\citeyear{kim2023crn})
    & RC & ResNet50 & $256\times704$&  56.0 & 49.0 &49.0\\
    RCBEVDet (\citeyear{lin2024rcbevdet}) & RC & ResNet50 &  $256\times704$ & 56.8 & 45.3&47.0\\
    \rowcolor{gray!20} RCTrans(ours) & RC & ResNet50 &  $256\times704$ & 58.6& 50.9 &52.2\\
    \hline
    RCBEV4d (\citeyear{huang2022bevdet4d})
    & RC & Swin-T & $256\times704$ & 49.7 & 38.1 & - \\
    RCBEVDet (\citeyear{lin2024rcbevdet}) & RC & Swin-T &  $256\times704$ & 56.2 & 49.6 &54.9\\
    \rowcolor{gray!20} RCTrans(ours) & RC & Swin-T &  $256\times704$ & \textbf{59.4}&\textbf{52.0}&60.5\\
    \hline
\end{tabular}
\end{center}
\caption{
    3D object detection performance comparison on nuScenes \texttt{val} set with different image backbones. 
     ‘C’ and ‘R’ represent camera and radar, respectively.}
\label{table1}
\end{table*}

\subsection{Training Head and Task Loss}

For the 3D object prediction head, we follow PETR \cite{liu2022petr} to use the FFN to regress each attribute of the object and conduct deep supervision. For the 3D object tracking, we follow CenterPoint \cite{yin2021center}  to use the 
velocity-based greedy matching algorithm. This is an offline reasoning process and does not require additional training. For the supervision of the model, we adopt focal loss \cite{lin2017focal} for classification and L1 loss for 3D bounding box regression. We utilize the Hungarian algorithm to assign the predicted objects and the ground-truths. The loss can be calculated by Eq. \ref{eq7}.
\begin{equation}\label{eq7}
L(o, \hat{o}) = \omega_1 L_{cls}(c, \hat{c}) + \omega_2 L_{reg}(b, \hat{b}),
\end{equation}
where $\omega_1$ and $\omega_2$ are the hyper-parameters to balance classification and regression loss. The variables $c$ and $b$ are the predicted classification and bounding box, while $\hat{c}$ and $\hat{b}$ are the corresponding ground truths. This kind of supervision is used for the predicted 3D boxes after each decoder layer.

\section{Experiments}
\subsection{Datasets and metrics}
Following previous approaches \cite{kim2023craft,kim2023crn,lin2024rcbevdet}, we conduct comprehensive experiments on a large-scale autonomous driving dataset for 3D radar-camera object detection, nuScenes \cite{caesar2020nuscenes}. This dataset has a total of 1000 scenes and is officially divided into 700/150/150 for training/validation/testing. For each frame, nuScenes has 6 images and 5 radar point clouds covering $360^{\circ}$. There are around 1.4 million annotated 3D bounding boxes for ten classes. We select nuScenes detection score (NDS) and mean average precision (mAP) as metrics to evaluate the 3D detection. 
For NDS, it is the weighted sum of mAP and other official predefined metrics, including Average Translation Error (ATE), Average Scale Error (ASE), Average Orientation Error (AOE), Average Velocity Error (AVE), Average Attribute Error (AAE). For 3D tracking, we use the official multi-object tracking accuracy (AMOTA), average multi-object tracking precision (AMOTP), false positive (FP), false negative (FN), and ID switches (IDS) as the metrics.
\subsection{Implementation Details}
We implement RCTrans based on StreamPETR \cite{wang2023exploring} and MMDetection3D \cite{mmdet3d2020} codebases. Following CRN \cite{kim2023crn}, we accumulate 
the information of 4 previous frames to the current frames. We use the object-centric temporal modeling proposed in StreamPETR to conduct temporal fusion. The number of decoder layers is set to 6 during training and 3 during inference.
The output of the last layer after pruning is inserted into the memory queue in temporal fusion.
We set the number of queries, memory queue, and propagated queries to 900, 512, and 128, respectively.  For radar, we accumulate 6 previous radar sweeps following
CRAFT \cite{kim2023craft} and set the maximum number of radar points to 2048. 
The size of radar BEV is set to 128$\times$128. 
We train our network for 90 epochs with a batch size of 32 on 8 NVIDIA A100 GPUs. The speed is evaluated on a single NVIDIA RTX3090 GPU. We conduct optimization based on AdamW with weight decay value $10^{-2}$, and the learning rate is adjusted by cycle policy with the initial value $4\times10^{-4}$.

\subsection{Performance Comparison} 

\begin{table*}
    
    \centering
    \begin{tabular}{c|c|c|ccccc|cc}
    \hline
    Method&Input&Backbone&mATE&mASE&mAOE&mAVE&mAAE&NDS$\uparrow$&mAP$\uparrow$\\
    \hline
    Radar-PointGNN (\citeyear{svenningsson2021radar})&R&-&1.024&0.859&0.897&1.020&0.931&3.4&0.5\\
    KPConvPillars (\citeyear{ulrich2022improved})&R&-&0.823&0.428&0.607&2.081&1.000&13.9&4.9\\
    \hline
    BEVFormer(\citeyear{li2022bevformer})&C&V2-99&0.582&0.256&0.375&0.378&0.126&56.9&48.1\\
    PETR-v2 (\citeyear{liu2023petrv2})&C&V2-99&0.561&0.243&0.361&0.343&0.120&58.2&49.0\\
    BEVDepth (\citeyear{li2023bevdepth})&C&ConvNeXt-B&0.445&0.243&0.352&0.347&0.127&60.9&52.0\\
    BEVStereo (\citeyear{li2023bevstereo})&C&ConvNeXt-B&0.431&0.246&0.358&0.357&0.138&61.0&52.5\\
    SOLOFusion (\citeyear{park2022time})&C&ConvNeXt-B&0.453&0.257&0.376&0.276&0.148&61.9&54.0\\
    SparseBEV (\citeyear{liu2023sparsebev})&C&V2-99&0.485&0.244&0.332&0.246&0.117&63.6&55.6\\
    StreamPETR (\citeyear{wang2023exploring})&C&V2-99&0.493&0.241&0.343&0.243&0.123&63.6&55.0\\
    \hline
    CenterFusion (\citeyear{nabati2021centerfusion})&RC&DLA34&0.631&0.261&0.516&0.615&0.115&44.9&32.6\\
    RCBEV (\citeyear{zhou2023bridging})&RC&Swin-T&0.484&0.257&0.587&0.702&0.140&48.6&40.6\\
    MVFusion (\citeyear{wu2023mvfusion})&RC&V2-99&0.569&0.246&0.379&0.781&0.128&51.7&45.3\\
    CRAFT (\citeyear{kim2023craft})&RC&DLA34&0.467&0.268&0.456&0.519&0.114&52.3&41.1\\
    CRN (\citeyear{kim2023crn})&RC&ConvNeXt-B&0.416&0.264&0.456&0.365&0.130&62.4&57.5\\
    RCBEVDet (\citeyear{lin2024rcbevdet})&RC&V2-99&0.390&0.234&0.362&0.259&0.113&63.9&55.0\\
    \hline
    RCTrans(ours)&RC&V2-99&0.459&0.245&0.392&0.198&0.121&\textbf{64.7}&\textbf{57.8}\\
    \hline
    \end{tabular}
    \caption{3D object detection performance comparison on nuScenes \texttt{test} set. The V2-99 
    \cite{lee2020centermask} backbone is pre-trained on external dataset DDAD \cite{guizilini20203d}.} 
    \label{table2}
\end{table*}

\begin{figure*}[h]
    \centering
    \includegraphics[scale=0.53]{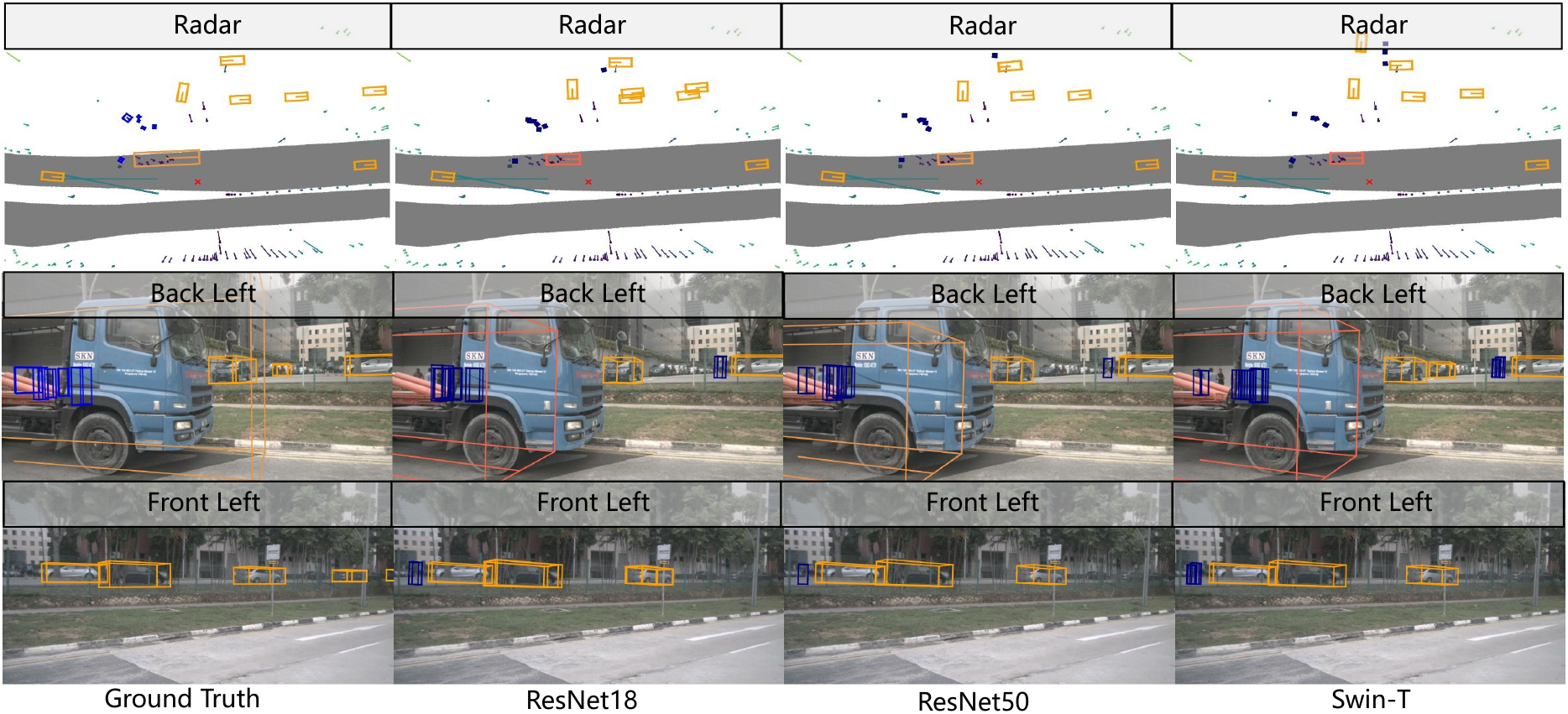}
    \caption{Qualitative results of RCTrans on nuScenes \texttt{val} set with different backbones. Due to space constraints, we only show the corresponding front left and back left images where objects are mainly distributed in this scene.} 
    \label{fig4}
\end{figure*}

\textbf{3D Object Detection.} 
We first compare our method with the state-of-the-art on nuScenes \texttt{val} set under different images backbones, including ResNet18 \cite{he2016deep}, ResNet50, and Swin-T \cite{liu2021swin}. As shown in Table \ref{table1}, our method achieves the best performance under different backbones with a slight increase in time consumption. For example, when using Swin-T as the backbone and setting the image size to 256$\times$704, compared to the SOTA radar-camera solution RCBEVDet, RCTrans improves NDS by 3.2\% and mAP by 2.4\%, while the latency increases around 5 ms. 
The great experimental results of multiple backbones demonstrate that RCTrans has excellent and strong adaptability, which is very beneficial for model deployment and migration in practical applications. What's more, RCTrans beats all the camera-only methods, including our camera stream baseline StreamPETR, which proves our method can effectively use radar information to supplement the detection results. We also compare our method with the state-of-the-art radar, camera, and radar-camera approaches on nuScenes \texttt{test} set. As shown in Table \ref{table2}, our proposed RCTrans outperforms all the competitive methods by achieving 64.7\% NDS and 57.8\% mAP.  Notably, RCTrans receives a giant performance improvement in mAVE, which proves that our model extracts useful velocity compensation information from radar data.
Qualitative results are shown in Fig. \ref{fig4}.

\begin{table}
    \centering
    \setlength{\tabcolsep}{0.4mm}
    \begin{tabular}{c|c|cc|ccc}
    \hline
    Method&Input&\small{AMOTA$\uparrow$}&\small{AMOTP$\downarrow$}&FP$\downarrow$&FN$\downarrow$&IDS$\downarrow$\\
    \hline
    DEFT(\citeyear{chaabane2021deft})&C&17.7&1.564&22163&60565&6901\\
    QD-3DT(\citeyear{hu2022monocular})&C&21.7&1.550&16495&60156&6856\\
    CC-3DT(\citeyear{fischer2022cc})&C&41.0&1.274&18114&42910&3334\\
    Sparse4D(\citeyear{lin2022sparse4d})&C&51.9&1.078&19626&32954&1090\\
    \hline
    CRN(\citeyear{kim2023crn})&RC&56.9&\textbf{0.809}&16822&41093&946\\
    RCTrans &RC&\textbf{59.6}&0.969&\textbf{14700}&\textbf{29491}&\textbf{654}\\
    \hline
    \end{tabular}
    \caption{3D object tracking performance comparison on nuScenes \texttt{test} set. We conduct experiments based on the V2-99 backbone.}
    \label{table3}
    \vspace{-0.5cm}
\end{table}

\textbf{3D Object Tracking.} We further prove the generalization ability of RCTrans on 3D object tracking task. As shown in Table \ref{table3}, we compare RCTrans with the existing camera and radar-camera tracking solutions on nuScenes \texttt{test} set. Overall, our approach yields the best results. Compared to CRN, our method significantly improves AMOTA, FP, FN, and IDS. The tracking results we compared are all obtained based on velocity-based closest distance matching in CenterPoint \cite{yin2021center}, 
so the improvement in tracking performance is mainly due to the more accurate prediction of the speed by our approach.

\subsection{Ablation Study}

\begin{table}[h]
    \centering
    \setlength{\tabcolsep}{0.9mm}
    \begin{tabular}{c|c|cc}
    \hline
    Component&Details&NDS&mAP\\
    \hline
    \multirow{4}{*}{Modality}&radar$^{\triangle}$&12.3&1.6\\
    &camera$^{\bigtriangledown}$&49.3&38.9\\            &radar+camera&51.7($\uparrow$2.4)&43.0($\uparrow$4.1)\\
    &radar+camera$^{\diamondsuit}$&55.1$(\uparrow$3.4)&46.6($\uparrow$3.6)\\
    \hline
     \multirow{3}{*}{\shortstack{RDE}}&+U-structure&54.2($\uparrow$2.5)&44.6($\uparrow$1.6)\\
    &+skip connection&56.3($\uparrow$2.1)&47.4($\uparrow$2.8)\\
    &+self-attention&56.4($\uparrow$0.1)&48.2($\uparrow$0.8)\\
    \hline
    \multirow{3}{*}{\shortstack{PSD}}&+position update&57.2($\uparrow$0.8)&49.7($\uparrow$1.5)\\
    &+sequential structure&58.4($\uparrow$1.2)&50.3($\uparrow$0.6)\\
    &+pruning strategy &58.6($\uparrow$0.2)&50.9($\uparrow$0.6)\\
    \hline
    \end{tabular}
    \caption{Ablation study on nuScenes \texttt{val} set. ${\triangle}$: the normal radar stream does not use any encoder on the BEV feature. ${\bigtriangledown}$: the camera stream is modified based on StreamPETR. We simplify it by removing the spatial alignment module and change the query position embedding. ${\diamondsuit}$: using SECOND as the BEV encoder.}
    \label{table4}
\end{table}

We conduct the ablation study on nuScenes \texttt{val} set to demonstrate the effectiveness of each component. All of our experiments are conducted under the condition of using ResNet50 as the backbone and setting the image size to 256$\times$704. As shown in Table \ref{table4},  each component can consistently improve performance. Compared to the single-modality stream, using multi-modality inputs can significantly achieve improvement. 
Compared to commonly used BEV encoders, such as SECOND \cite{yan2018second}, using our proposed Radar Dense Encoder (RDE) can obtain 1.3$\%$ NDS and 1.6$\%$ mAP improvement. 
\begin{table}[t!]
    \centering
    \setlength{\tabcolsep}{0.8mm}
    \begin{tabular}{c|c|ccc}
    \hline
    Training layer&Inference layer&NDS&mAP&latency(ms)\\
    \hline
    6&6&58.4&50.3&69.9\\
    6&5&58.5&50.5&64.0\\
    6&4&\textbf{58.7}&\textbf{51.0}&58.1\\
    6&3&58.6&50.9&\textbf{52.2}\\
    3&3&56.8&49.2&52.2\\
    \hline
    \end{tabular}
    \caption{Ablation study of pruning training strategy.}
    \label{table5}
    \vspace{-0.5cm}
\end{table}
What's more, the Pruning Sequential Decoder (PSD) increases NDS by 2.2$\%$ and mAP by 2.7$\%$. We also conduct experiments on the pruning training strategy as shown in Table \ref{table5}. We find that setting the number of decoders to 6 during training and only using the first 3 layers during inference results in great performance, reducing inference time by 17.7ms while increasing NDS by 0.2\%. The reason for this phenomenon may be that recalculating the position embedding after each layer can quickly obtain accurate alignment and faster converge but make some queries focus on the same region. Moreover, fusing different modal information in each layer by sequential structure will result in a large total number of transformers in the decoder. Reducing the number of decoder layers during inference can speed up the inference procedure while the performance is not reduced.

\subsection{Robustness Analysis}

\begin{table}[!t]

\begin{center}
\setlength{\tabcolsep}{1.6mm}
\begin{tabular}{l|c|c|cccc}
\hline
& \multirow{2}{*}{Input} & \multirow{2}{*}{Drop} & \multicolumn{4}{c}{\# of view drops} \\
&  &  & 0 & 1 & 3 & 6 \\
\hline
BEVDepth  &  C  & C & 49.4 & 41.1 & 24.2 & 0 \\
CenterPoint &  R  & R & 30.6 & 25.3 & 14.9 & 0 \\
\hline
\multirow{2}{*}{BEVFusion} & \multirow{2}{*}{C+R} & C & \multirow{2}{*}{63.9} 
& 58.5 & 45.7 & {14.3} \\
& & R & & 59.9 & 50.9 & 34.4 \\
\hline
\multirow{2}{*}{CRN} & \multirow{2}{*}{C+R} & C & \multirow{2}{*}{\shortstack{{68.8}}} 
& {62.4} & {48.9} & 12.8 \\
& & R & & {64.3} & {57.0} & {43.8} \\
\hline
\multirow{2}{*}{RCBEVDet} & \multirow{2}{*}{C+R} & C & \multirow{2}{*}{\shortstack{72.5}} 
& 66.9 & 53.5 & 16.5 \\
& & R & & 71.6 & 66.1 & 62.1 \\
\hline
\multirow{2}{*}{RCTrans} & \multirow{2}{*}{C+R} & C & \multirow{2}{*}{\shortstack{\textbf{73.4}}} 
& \textbf{71.9} & \textbf{68.8} & \textbf{20.3} \\
& & R & & \textbf{72.2} & \textbf{71.6} & \textbf{62.5} \\
\hline
\end{tabular}
\end{center}
\caption{
Robustness analysis via car class mAP. When drops$=$6, it means that the inputs contain single modality.}
\label{table6}
\vspace{-0.5cm}
\end{table}

To simulate the sensor malfunction in real autonomous driving scenarios, we randomly drop images or radar inputs and evaluate the results. For a fair comparison, following CRN \cite{kim2023crn} and RCBEVDet \cite{lin2024rcbevdet}, we re-train the
methods with data drop augmentation proposed in Autoalignv2 \cite{chen2022deformable}, which randomly drops 3 image inputs during training.
During the test process, the number of the dropped sensors is set to 1, 3, and 6.
The removed inputs will be replaced by a fully zero tensor of the same shape. Car class mAP is used to evaluate the robustness of the methods. As shown in Table \ref{table6}, RCTrans outperforms all the existing methods under different sensor failure cases. Specifically, compared to RCBEVDet, RCTrans improves mAP by 5.0$\%$, 15.3$\%$, and 3.8$\%$ under these three camera failure conditions. Notably, when the number of dropped inputs is 3, RCTrans can still get the results close to the condition with null or 1 dropped input, which is an important improvement over past methods.
The above phenomenon shows that our model can achieve a more stable effect under the sensor failure cases. 

\section{Conclusion}

In this paper, we propose a Radar-Camera Transformer (RCTrans), a query-based method that can effectively solve the fusion difficulties caused by the sparsity and noise of radar inputs. To solve the problem of sparsity, we introduce the Radar Dense encoder which uses a downsample-then-upsample architecture. 
It can adaptively fill the invalid BEV grids while maintaining the information of small objects.
To solve the problem of noise, we introduce the Pruning Sequential Decoder which uses a step-by-step strategy to gradually regress the objects. 
We also introduce the pruning training strategy for the decoder, which saves much time during inference while also preventing attention collapse.
Extensive experiments demonstrate the effectiveness and robustness of RCTrans. We believe RCTrans can be an effective baseline to inspire future research.

\section{Acknowledgments}
This work was supported in part by the Chinese National Natural Science Foundation Project 62206276, 62276254, U23B2054, and the InnoHK program.


\bibliography{aaai25}

\end{document}